%% file: main.tex
\documentclass[10pt,twocolumn,letterpaper]{article}

\usepackage[pagenumbers]{cvpr} % To force page numbers, e.g. for an arXiv version

\input{preamble}

\definecolor{cvprblue}{rgb}{0.21,0.49,0.74}
\usepackage[pagebackref,breaklinks,colorlinks,allcolors=cvprblue]{hyperref}
\usepackage{url}
\usepackage[utf8]{inputenc} % allow utf-8 input
\usepackage[T1]{fontenc}    % use 8-bit T1 fonts
\usepackage{booktabs}       % professional-quality tables
\usepackage{amsfonts}       % blackboard math symbols
\usepackage{nicefrac}       % compact symbols for 1/2, etc.
\usepackage{microtype}      % microtypography
\usepackage{multirow}
\usepackage{amsmath}
\usepackage{graphicx}
\usepackage{colortbl}
\usepackage{afterpage}
\usepackage{longtable}

\title{Medical SAM 2: Segment Medical Images as Video via Segment Anything Model 2}

\author{Jiayuan Zhu\\
University of Oxford\\
{\tt\small jiayuan.zhu@ieee.org}
\and
Abdullah Hamdi \\
University of Oxford\\
{\tt\small abdullah.hamdi@eng.ox.ac.uk}
\and
Yunli Qi \\
University of Oxford\\
{\tt\small yunli.qi@wolfson.ox.ac.uk}
\and
Yueming Jin \\
National University of Singapore\\
{\tt\small ymjin@nus.edu.sg}
\and
Junde Wu \\
University of Oxford\\
{\tt\small jundewu@ieee.org}
}

\begin{document}
\maketitle
\input{sec/0_abstract}

\input{sec/1_intro}

\input{sec/2_related}
\input{sec/3_method}

\input{sec/4_experiment}

\input{sec/5.results}

\input{sec/6_conclusion}
{
    \small
    \bibliographystyle{ieeenat_fullname}
    \bibliography{main}
}
\clearpage \clearpage
% \clearpage
% % WARNING: do not forget to delete the supplementary pages from your submission 
\appendix
\input{sec/X_suppl}

\end{document}

%% file: preamble.tex
\usepackage{amsmath,amsfonts,bm}

\definecolor{green}{RGB}{0,150,10}

\DeclareRobustCommand{\figLabel}{Figure~}
\DeclareRobustCommand{\eqLabel}[1]{{Eq (#1)}}

\DeclareRobustCommand{\mysection}[1]{\noindent\textbf{#1.}}
\DeclareRobustCommand{\supp}{\textit{Appendix}\xspace}

%% file: sec/0_abstract.tex
\begin{abstract}
Medical image segmentation plays a pivotal role in clinical diagnostics and treatment planning, yet existing models often face challenges in generalization and in handling both 2D and 3D data uniformly. In this paper, we introduce Medical SAM 2 (MedSAM-2), a generalized auto-tracking model for universal 2D and 3D medical image segmentation. The core concept is to leverage the Segment Anything Model 2 (SAM2) pipeline to treat all 2D and 3D medical segmentation tasks as a video object tracking problem. To put it into practice, we propose a novel \emph{self-sorting memory bank} mechanism that dynamically selects informative embeddings based on confidence and dissimilarity, regardless of temporal order. This mechanism not only significantly improves performance in 3D medical image segmentation but also unlocks a \emph{One-Prompt Segmentation} capability for 2D images, allowing segmentation across multiple images from a single prompt without temporal relationships. We evaluated MedSAM-2 on five 2D tasks and nine 3D tasks, including white blood cells, optic cups, retinal vessels, mandibles, coronary arteries, kidney tumors, liver tumors, breast cancer, nasopharynx cancer, vestibular schwannoma, mediastinal lymph nodules, cerebral artery, inferior alveolar nerve, and abdominal organs, comparing it against state-of-the-art (SOTA) models in task-tailored, general and interactive segmentation settings. Our findings demonstrate that MedSAM-2 surpasses a wide range of existing models and updates new SOTA on several benchmarks. The code is released on the project page: \href{https://supermedintel.github.io/Medical-SAM2/}{https://supermedintel.github.io/Medical-SAM2/}.

% In this paper, we introduce Medical SAM 2 (MedSAM-2), an advanced segmentation model that utilizes the SAM 2 framework to address both 2D and 3D medical image segmentation tasks. By adopting the philosophy of taking medical images as videos, MedSAM-2 not only applies to 3D medical images but also unlocks new One-prompt Segmentation capability. That allows users to provide a prompt for just one or a specific image targeting an object, after which the model can autonomously segment the same type of object in all subsequent images, regardless of temporal relationships between the images. We evaluated MedSAM-2 across a variety of medical imaging modalities, including abdominal organs, optic discs, brain tumors, thyroid nodules, and skin lesions, comparing it against state-of-the-art models in both traditional and interactive segmentation settings. Our findings show that MedSAM-2 not only surpasses existing models in performance but also exhibits superior generalization across a range of medical image segmentation tasks. Our code is attached and will be released upon publication.
% Our code will be released at: https://github.com/MedicineToken/Medical-SAM2
\end{abstract}

%% file: sec/1_intro.tex
\vspace{-5pt}
\section{Introduction}
\label{sec:intro}

\begin{figure}[t]
    \begin{center}
\includegraphics[page=1,trim={0 0 18cm 7cm},clip,width=0.95\linewidth]{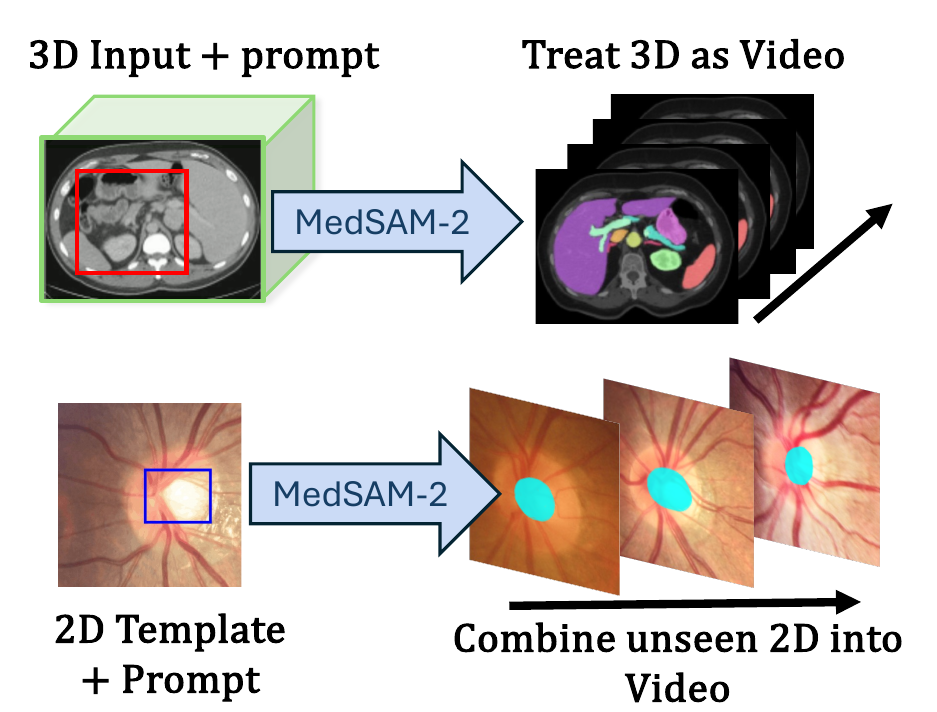}    
\vspace{-10pt}
\end{center}
    \caption{\textbf{Segmentation Capabilities of MedSAM-2}. When provided with a prompt in one 3D slice, MedSAM-2 can segment all later spatial-temporal 3D frames. When given a prompt in one 2D image, MedSAM-2 can accurately segment other 2D images that are not temporally related using the same criteria, which is an emergence of One-prompt Segmentation capability. }\label{fig:cover}
\vspace{-10pt}
\end{figure}

\begin{figure*}[t]
    \begin{center}
    %\framebox[4.0in]{$\;$}
\includegraphics[page=2,trim={0 0 0 8cm},clip,width=0.98\linewidth]{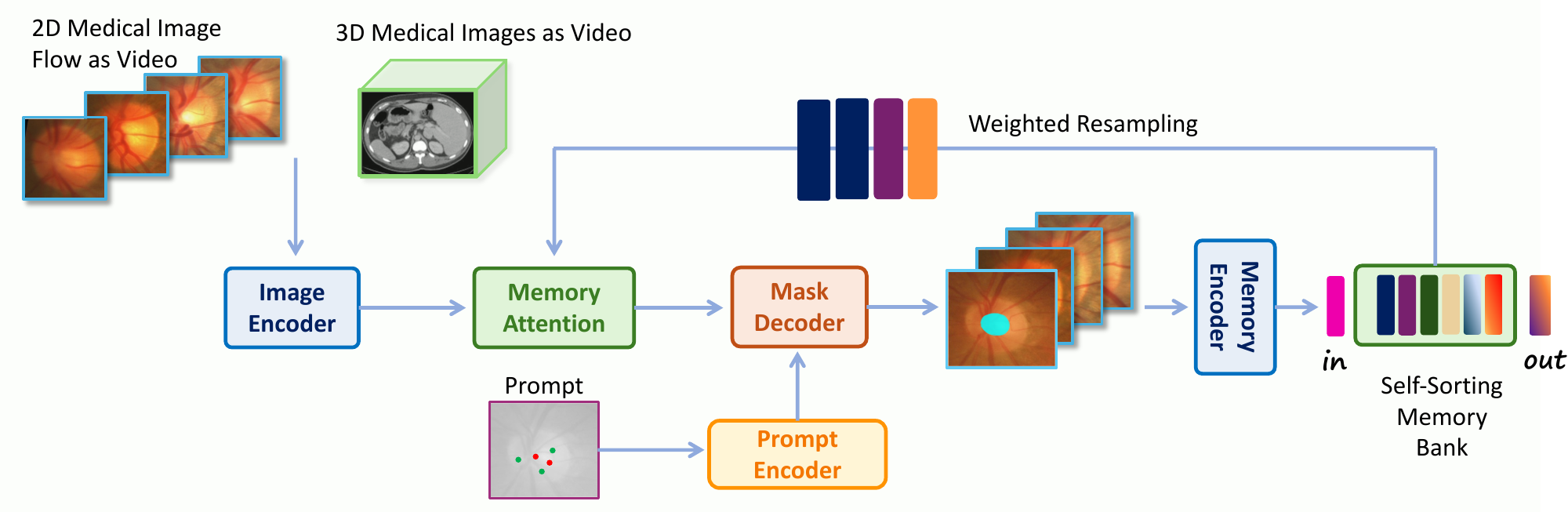}    \end{center}
\vspace{-10pt}
    \caption{\textbf{MedSAM-2 Framework.} Building on the SAM 2 framework, we propose treating 3D medical images and 2D medical image flows as videos to facilitate memory-enhanced medical image segmentation. This approach not only improves performance in 3D medical image segmentation but also unlocks One-Prompt Segmentation capability for 2D medical image flows. This is achieved by incorporating our proposed Self-Sorting Memory Bank, which selects the most confident embeddings based on the confidence predictions ($\alpha$, $\beta$, $\gamma$) from the mask decoder.}\label{fig1}
\vspace{-10pt}
\end{figure*}

Artificial intelligence has significantly transformed various industries, and healthcare is poised for a substantial revolution driven by advancements in medical image understanding \cite{wei2022emergent, kirillov2023segment, wang2023seggpt, wang2023images,xdiffusion}. Medical image segmentation, which involves partitioning images into meaningful regions, is crucial for applications like diagnosis, treatment planning, and image-guided surgery \cite{wu2022medsegdiff, wu2023medical, wu2024medsegdiffv2}. Despite the progress made with automated segmentation methods using deep learning models such as convolutional neural networks (CNNs) and vision transformers (ViTs), significant challenges remain \cite{ronneberger2015u, alexey2020image}. One primary issue is model generalization; models trained on specific targets like certain organs or tissues often struggle to adapt to other targets or modalities. Additionally, many deep learning architectures are designed for 2D images, whereas medical imaging data frequently exist in 3D formats (e.g., CT, MRI), creating a gap when applying these models to 3D data \cite{3dunet, milletari2016v}.

Recent developments in promptable segmentation models, particularly the Segment Anything Model (SAM) \cite{kirillov2023segment} and its enhanced version SAM 2 \cite{ravi2024sam2}, have shown promise in addressing some challenges. SAM has demonstrated remarkable zero-shot capabilities in image segmentation tasks by leveraging user-provided prompts to segment objects without prior training on specific targets. However, this approach requires user interaction for each image, which can be labor-intensive and impractical in clinical settings where large volumes of data are common \cite{MedSAM}. SAM 2 extends SAM's capabilities to  videos, introducing real-time object tracking with reduced user interaction time. Yet, it still relies on temporal relationships between frames, limiting its applicability to unordered medical images and failing to fully address the generalization challenges in medical image segmentation.

In this work, we introduce MedSAM-2, a generalized auto-tracking model for universal medical image segmentation. MedSAM-2 tackles these challenges by treating medical images as videos and incorporating a novel \textbf{self-sorting memory bank}.
% In this work, we introduce MedSAM-2, a generalized foundation segmentation model for universal medical imaging that addresses these challenges by treating medical images as videos and incorporating a novel \textbf{self-sorting memory bank}. 
This mechanism dynamically selects informative embeddings based on confidence and dissimilarity, allowing the model to handle unordered medical images effectively. By rethinking the memory mechanism in SAM 2, MedSAM-2 not only improves performance in 3D medical image segmentation but also unlocks the \emph{One-Prompt Segmentation} capability \cite{Wu_2024_CVPR} for 2D medical images. This capability enables the model to generalize from a single prompt to segment across multiple images without temporal relationships, significantly reducing user interaction and enhancing convenience for clinicians. 
% MedSAM-2 achieves this by increasing the entropy and mutual information in the memory bank, leading to better generalization and robustness in segmentation tasks.

We evaluated MedSAM-2 across 14 different benchmarks, encompassing 25 distinct tasks for validation. Compared with previous fully-supervised segmentation models and SAM-based interactive models, MedSAM-2 demonstrated superior performance across all tested methods and achieved state-of-the-art results in both 2D and 3D medical image segmentation tasks. Specifically, under the one-prompt segmentation setting, MedSAM-2 outperformed previous foundation segmentation models, thereby showcasing its exceptional generalization capabilities. Our contributions can be summarized as follows:

\mysection{Contributions} 
\textbf{(i)} We introduce MedSAM-2, the first SAM-2-based generalized auto-tracking model for universal medical image segmentation, capable of uniformly handling both 2D and 3D medical imaging tasks with minimal user intervention.
\textbf{(ii)} We propose a novel \textit{self-sorting memory bank} mechanism that dynamically selects informative embeddings based on confidence and dissimilarity, enhancing the model's ability to handle unordered medical images and improves generalization.
\textbf{(iii)} We evaluate MedSAM-2 across 15 different benchmarks, including 25 distinct tasks, demonstrating superior performance compared to previous fully-supervised and SAM-based interactive models.

%% file: sec/2_related.tex
\section{Related Works}
\label{sec:related}
\mysection{Medical Image Segmentation}
Traditionally, medical image segmentation models have been task-specific, designed and optimized for particular targets like specific organs or tissues \cite{chen2021transunet, chen2023transattunet}. These task-tailored models leverage the unique characteristics of each task to achieve high performance. For instance, uncertain-aware modules have been utilized to handle the ambiguity in optic cup segmentation in fundus images \cite{ji2021learning}
%, while dynamic convolution kernels have been proposed to adapt to the complex structures of the pancreas in abdominal MRI images \cite{hatamizadeh2022swin}.
However, the reliance on task-specific models presents significant challenges. Designing and training a unique model for each segmentation task is labor-intensive and time-consuming. Moreover, many deep learning architectures are designed for 2D images, whereas medical imaging data often exist in 3D formats (e.g., CT, MRI), creating a gap when applying these models to 3D data \cite{3dunet, milletari2016v,mvtn,hamdi2023voint,Hamdi2024ijcv}.
To address these limitations, there has been growing interest in developing generalized medical image segmentation models capable of handling multiple tasks and modalities \cite{hatamizadeh2022unetr, wu2022medsegdiff, wu2024medsegdiffv2}. These models aim to generalize across different targets without the need for task-specific adaptations. However, achieving robust generalization remains a significant challenge due to the diverse nature of medical images, which can vary greatly in appearance, resolution, and anatomical structures. On the other hand, our MedSAM-2 is a generalized model that tackles multiple domains and can be used for both 2D and 3D medical image segmentation. 

\mysection{Prompting Segment Anything Models (SAMs)}
The introduction of the Segment Anything Model (SAM) \cite{kirillov2023segment} marked a significant advancement in the field of image segmentation. SAM leverages user-provided prompts to segment objects in images without prior training on specific targets, demonstrating remarkable zero-shot capabilities. In medical imaging, early applications of SAM involved fine-tuning the model to adapt to different medical segmentation tasks \cite{roy2023sam, deng2023sam, cheng2023sam, wu2023medical}. However, these approaches still required user interaction for each image, which can be impractical in clinical settings with large volumes of data.
To reduce the reliance on extensive user prompting, researchers have explored few-shot and zero-shot segmentation methods \cite{ding2023few, wu2020leveraging, li2023prototypical, ouyang2020self}, enabling adaptation to new tasks with minimal annotated samples. For example, UniSeg \cite{butoi2023universeg} requires only a few annotated samples to process an entire unseen segmentation task during inference. One-Prompt Segmentation \cite{Wu_2024_CVPR} further combines SAM's interactive setting with zero-shot segmentation, requiring only one visual prompt for a template sample to segment similar targets in subsequent samples without retraining. Nonetheless, these models still primarily focus on 2D medical images and are not yet tailored for the unique requirements of 3D medical imaging. Our MedSAM-2 uses a self-sorting memory bank allowing the model to generalize better on unsorted medical images while leveraging the context-rich video pretraining of SAM 2 with minimal prompting requirements. 

%% file: sec/3_method.tex
\section{Method}
\label{sec:method}

We introduce MedSAM-2, an advanced segmentation model based on SAM 2~\cite{ravi2024sam2}, tailored for medical image segmentation tasks.

\subsection{Preliminaries on Segment Anything Model (SAM 2)}
\label{sec:preliminaries}

SAM 2~\cite{ravi2024sam2} is a promptable visual segmentation model designed for image and video tasks. Given an input sequence of frames or images $\mathbf{X} = \{\mathbf{x}_t\}_{t=1}^T$ and optional prompts $\mathbf{P} = \{\mathbf{P}_t\}_{t=1}^T$, the model predicts segmentation masks $\mathbf{Y} = \{\mathbf{y}_t\}_{t=1}^T$ for each frame $\mathbf{x}_t$. The architecture comprises an image encoder $\mathcal{E}_{\text{img}}$ that encodes each frame $\mathbf{x}_t$ into a feature embedding $\mathbf{F}_t = \mathcal{E}_{\text{img}}(\mathbf{x}_t)$; a prompt encoder $\mathcal{E}_{\text{prompt}}$ that processes user prompts $\mathbf{P}_t$, generating embeddings $\mathbf{Q}_t = \mathcal{E}_{\text{prompt}}(\mathbf{P}_t)$; a memory bank $\mathcal{M}_t$ that stores $K$ past embeddings $\mathbf{E}_i$ before frame $\mathbf{x}_t$; a memory attention mechanism $\mathcal{A}$ that combines $\mathbf{F}_t$, $\mathcal{M}_t$, and $\mathbf{Q}_t$; and a mask decoder $\mathcal{D}$ that predicts the segmentation mask $\mathbf{y}_t$. Mathematically, the segmentation process can be formulated:
\begin{equation} \label{eq:sam2}
\begin{aligned} 
    \mathbf{y}_t &= \mathcal{D}\left( \mathcal{A}\left( \mathbf{F}_t, \mathcal{M}_t, \mathbf{Q}_t \right) \right), \quad \text{for } t = 1, \ldots, T, \\
    \mathcal{M}_t &= \left\{ \mathbf{E}_i \mid i \in \left\{ \max(j,0) \right\}_{j=t-K-1}^{t-1} \right\}, 
\end{aligned}
\end{equation}

\subsection{MedSAM-2: Self-Sorting SAM2 for Medical Imaging}
\label{sec:medsam2}

Although SAM2 has been highly successful with natural images, directly applying it to medical images is not straightforward. In medical imaging, the order of slices or images may not be meaningful due to varying acquisition protocols and orientations. Moreover, 2D medical images are often unordered, and each orientation in 3D imaging can be considered as an independent sequence to be integrated with different order. To address this, we propose a \textbf{self-sorting memory bank} $\mathcal{M}_t^{\text{sort}}$ that dynamically selects and retains the most informative embeddings, rather than simply using the most recent $K$ frames as in SAM 2~\cite{ravi2024sam2}.

\mysection{Memory Bank Update with Confidence and Dissimilarity}
\label{sec:memory_update}
At each time step $t$, we update the self-sorting memory bank $\mathcal{M}_t^{\text{sort}}$ based on $\mathcal{M}_{t-1}^{\text{sort}}$ and the embedding $\mathbf{E}_{t-1}$ of the previous frame.
First, the model predicts the segmentation mask $\mathbf{y}_{t-1}$ and computes IOU confidence score $c_{t-1}$ for frame $t-1$, estimated by the model itself.
If the confidence score satisfies $c_{t-1} \geq c_{\text{thresh}}$, we consider adding $\mathbf{E}_{t-1}$ to the memory bank. We form a candidate set $\mathcal{C} = \mathcal{M}_{t-1}^{\text{sort}} \cup \{ \mathbf{E}_{t-1} \}$.
To maintain diversity, we compute the total dissimilarity for each embedding in $\mathcal{C}$:
\begin{equation}
    D_i = \sum_{\substack{\mathbf{E}_j \in \mathcal{C} \\ j \neq i}} 1 - \text{sim}(\mathbf{E}_i, \mathbf{E}_j), \quad \forall \mathbf{E}_i \in \mathcal{C} = \mathcal{M}_{t-1}^{\text{sort}} \cup \{ \mathbf{E}_{t-1} \},
\end{equation}
where $\text{sim}(\cdot,\cdot)$ is a similarity function (e.g., cosine similarity).
We then select the top $K$ embeddings with the highest total dissimilarity to form the updated memory bank:
\begin{equation} \label{eq:memory_update}
    \mathcal{M}_t^{\text{sort}} = \operatorname*{TopK}_{\mathbf{E}_i \in \mathcal{C}} (D_i).
\end{equation}
If the confidence condition is not met ($c_{t-1} < c_{\text{thresh}}$), we keep the memory bank unchanged 
 $ \mathcal{M}_t^{\text{sort}} = \mathcal{M}_{t-1}^{\text{sort}}.$

\mysection{Resampling the Memory Bank}
\label{sec:resampling_memory_bank}
Before computing the attention for frame $t$, we resample the memory bank to emphasize embeddings similar to the current embedding $\mathbf{F}_t$, enhancing relevance. This is achieved by assigning higher selection probabilities to embeddings more similar to $\mathbf{F}_t$.
We calculate the similarity scores between $\mathbf{F}_t$ and each embedding $\mathbf{E}_i$ in the memory bank $\mathcal{M}_t^{\text{sort}}$ using a similarity function $\text{sim}(\cdot,\cdot)$ (e.g., cosine similarity):
\begin{equation}
    p_{i,t} = \frac{\text{sim}(\mathbf{F}_t, \mathbf{E}_i)}{\sum_{\mathbf{E}_j \in \mathcal{M}_t^{\text{sort}}} \text{sim}(\mathbf{F}_t, \mathbf{E}_j)}, \quad \forall \mathbf{E}_i \in \mathcal{M}_t^{\text{sort}}.
\end{equation}

Using the probability distribution $\{ p_{i,t} \}$, we perform resampling with replacement to create the importance-weighted memory bank $\tilde{\mathcal{M}}_t^{\text{sort}}$. Specifically, we sample $K$ embeddings from $\mathcal{M}_t^{\text{sort}}$, where each embedding $\mathbf{E}_i$ is selected independently with probability $p_{i,t}$:
\begin{equation} \label{eq:sortmemory}
    \tilde{\mathcal{M}}_t^{\text{sort}} = \left\{ \mathbf{E}_{i_k} \mid i_k \sim \text{Categorical}(\{ p_{i,t} \}), \; k = 1, \ldots, K \right\}.
\end{equation}

This resampling process effectively prioritizes embeddings more similar to current embedding $\mathbf{F}_t$, enhancing the relevance of the memory bank in the attention mechanism.

\mysection{MedSAM-2 Pipeline}
The segmentation process in MedSAM-2 incorporates the self-sorting memory bank and resampled embeddings into SAM 2. With the fixed prompt $\mathbf{P}_1$ from the first frame, we modify \eqref{eq:sam2} as:
\begin{equation} \label{eq:medsam2}
    \mathbf{y}_t = \mathcal{D}\left( \mathcal{A}\left( \mathbf{F}_t, \tilde{\mathcal{M}}_t^{\text{sort}}, \mathbf{Q}_1 \right) \right), \quad \text{for } t = 1, \ldots, T,
\end{equation}
where $\mathbf{F}_t = \mathcal{E}_{\text{img}}(\mathbf{x}_t)$, $\mathbf{Q}_1 = \mathcal{E}_{\text{prompt}}(\mathbf{P}_1)$, and $\tilde{\mathcal{M}}_t^{\text{sort}}$ is the resampled memory bank from \eqLabel{\ref{eq:sortmemory}}.
This modification allows MedSAM-2 to handle unordered medical images effectively, leveraging informative and relevant embeddings for segmentation, thus enhancing performance in both 2D and 3D medical imaging tasks after training with standard segmentation loss~\cite{MedSAM}.

\mysection{Self-Sorting Works because of Entropy and Mutual Information} \label{sec:self-sorting}
By utilizing the self-sorting memory bank, we ensure that the memory bank contains the most reliable and informative embeddings, regardless of their temporal order. This self-sorting mechanism not only handles the unordered nature of medical images but also forces the extracted features to have higher entropy due to the increased randomness introduced by the "learned shuffle`` based on confidence rather than inherent temporal order. This increased entropy coincides with an increase in the mutual information between the memory bank features and the output, improving the robustness and generalization of the model according to principle of maximum entropy \cite{jaynes1957information}. Consequently, the model is better equipped to handle unordered medical images. We show mathematically in \supp how entropy-increase and mutual information by self-sorting improve learning generalization.   
% \A{I might add a toy experiment here fitting an MLP that self-sort vs unsorted input to see how empirically on a toy experiment this helps}

\subsection{Unified Approach for 2D and 3D Images}
\label{sec:unified}
MedSAM-2 leverages a self-sorting memory bank to improve robustness and effectively utilize context in both 2D and 3D medical imaging segmentation tasks. This unified framework allows MedSAM-2 to perform effectively across diverse medical imaging scenarios, unlocking the 'One-Prompt Segmentation' capability for 2D medical images and improving performance in 3D segmentation.

\mysection{MedSAM-2 for 3D Medical Imaging} For 3D medical images, such as MRI or CT scans represented as volumes $\mathbf{V} \in \mathbb{R}^{H \times W \times D}$, we treat the volume as a sequence of 2D slices along various orientations, similar to frames in a video. We define six orientations for processing the 3D volume:

\begin{enumerate}
    \item \textbf{Axial}: $\mathbf{X}^{(1)} = \{ \mathbf{x}_t = \mathbf{V}(:, :, t) \}_{t=1}^D$.
    \item \textbf{Coronal}: $\mathbf{X}^{(2)} = \{ \mathbf{x}_t = \mathbf{V}(:, t, :) \}_{t=1}^H$.
    \item \textbf{Sagittal}: $\mathbf{X}^{(3)} = \{ \mathbf{x}_t = \mathbf{V}(t, :, :) \}_{t=1}^W$.
    \item \textbf{Reverse Axial}: $\mathbf{X}^{(4)} = \{ \mathbf{x}_t = \mathbf{V}(:, :, D - t + 1) \}_{t=1}^D$.
    \item \textbf{Reverse Coronal}: $\mathbf{X}^{(5)} = \{ \mathbf{x}_t = \mathbf{V}(:, H - t + 1, :) \}_{t=1}^H$.
    \item \textbf{Reverse Sagittal}: $\mathbf{X}^{(6)} = \{ \mathbf{x}_t = \mathbf{V}(W - t + 1, :, :) \}_{t=1}^W$.
\end{enumerate}

Processing the volume with all orientations combined $\mathbf{X} = \bigcup_{o=1}^{6} \mathbf{X}^{(o)}$ exposes the model to diverse anatomical perspectives, enhancing its ability to generalize and capture anisotropic structures. However, the best \textit{order} of picking these directions is still \textit{unknown}. Hence, our self-sorting memory bank order embeddings from different orientations based on the mean direction features and their confidences as before. This allows our MedSAM-2 model to \textit{jointly} capture the 3D context and reap the benefits of the self-sorting mechanism.

During inference, the model processes the input data in multiple orientations with combined input $\mathbf{X}$, obtaining segmentation predictions where the final output for the 3D volume is obtained by aggregating these predictions:
$\mathbf{Y}_{\text{3D}} = \operatorname{Aggregate}\left( \{ \mathbf{Y}^{(o)} \}_{o=1}^6 \right),
$
 where $\operatorname{Aggregate}$ is a function such as averaging or majority voting applied pixel-wise.

\mysection{MedSAM-2 for One-prompt 2D Segmentation}  For 2D medical images, which may consist of independent slices or images lacking temporal connections, we treat sets of images as pseudo-video sequences. By processing them sequentially using MedSAM-2's memory mechanism, we achieve a \textit{One-Prompt Segmentation} capability \cite{Wu_2024_CVPR}, where providing a prompt on a single image template ($\mathbf{X}_1,\mathbf{P}_1$) allows the model to propagate the segmentation across the entire set. Our MedSAM-2 approach leverages the self-sorting memory bank to associate the prompt more closely with intrinsic features in each frame, improving generalization and efficiency.
This ability of one-prompt-segmentation is less restrictive than the universal interactive video object segmentation (iVOS) \cite{MedSAM}, where its target is to learn a universal function for any single input image $\mathbf{x}_i$ and prompt $\mathbf{P}_i$, to predict the output mask $\mathbf{y}_i$.

%% file: sec/4_experiment.tex
\begin{table*}[h]
\centering
 
\caption{\textbf{3D Medical Images Segmentation Performance}. We show the comparison of MedSAM-2 with SOTA segmentation methods over BTCV dataset \cite{fang2020multi} evaluated by Dice Score (\%). Task-tailored models, interactive generalized models, auto-tracking generalized models are marked in yellow, green, blue.}
\vspace{-5pt}
\resizebox{\linewidth}{!}{%
\begin{tabular}{c|cccccccccccc|c}
\hline
Model                                                           & Spleen & R.Kid & L.Kid & Gall. & Eso.  & Liver & Stom.  & Aorta & IVC  &Veins & Panc. & AG  & Ave  \\ \hline
\rowcolor{yellow!35!white!40}
TransUNet                                                       & 0.952                        & 0.927 & 0.929 & 0.662 & 0.757 & 0.969  & 0.889 & 0.920  & 0.833 & 0.791       & 0.775     & 0.637 & 0.838 \\
\rowcolor{yellow!35!white!40}
UNetr                                                           & 0.968                        & 0.924 & 0.941 & 0.750 & 0.766 & 0.971  & 0.913 & 0.890  & 0.847 & 0.788       & 0.767     & 0.741 & 0.856 \\
\rowcolor{yellow!35!white!40}
Swin-UNetr                                                      & 0.971                        & 0.936 & 0.943 & 0.794 & 0.773 & 0.975  & 0.921 & 0.892  & 0.853 & 0.812       & 0.794     & 0.765 &  0.869    \\
\rowcolor{yellow!35!white!40}
nnUNet                                                          & 0.942                        & 0.894 & 0.910 & 0.704 & 0.723 & 0.948  & 0.824 & 0.877  & 0.782 & 0.720       & 0.680     & 0.616 & 0.802 \\
\rowcolor{yellow!35!white!40}
EnsDiff                                                         & 0.938                        & 0.931 & 0.924 & 0.772 & 0.771 & 0.967  & 0.910 & 0.869  & 0.851 & 0.802       & 0.771     & 0.745 & 0.854      \\
\rowcolor{yellow!35!white!40}
SegDiff                                                 & 0.954                        & 0.932 & 0.926 & 0.738 & 0.763 & 0.953  & 0.927 & 0.846  & 0.833 & 0.796       & 0.782     & 0.723 & 0.847     \\
\rowcolor{yellow!35!white!40}
MedSegDiff                                                      & 0.973                        & 0.930 & 0.955 & 0.812 & 0.815 & 0.973  & 0.924 & 0.907  & 0.868 & 0.825       & 0.788     & 0.779 & 0.879     \\ \hline
\rowcolor{green!45!white!10}
SAM & 0.518  & 0.686 & 0.791 & 0.543 & 0.584 & 0.461  & 0.562 & 0.612  & 0.402 & 0.553       & 0.511     & 0.354 & 0.548     \\
\rowcolor{green!45!white!10}
MedSAM   & 0.751  & 0.814 & 0.885 & 0.766 & 0.721 & 0.901  & 0.855 & 0.872  & 0.746 & 0.771       & 0.760     & 0.705 & 0.803        \\
\rowcolor{green!45!white!10}
SAM-U  & 0.868 & 0.776 & 0.834 & 0.690 & 0.710 & 0.922 & 0.805 & 0.863 & 0.844 & 0.782 & 0.611 & 0.780 & 0.790 \\
\rowcolor{green!45!white!10}
SAM-Med3D & 0.873 & 0.884 & 0.932 & 0.795 & 0.790 & 0.943 & 0.889 & 0.872 & 0.796 & 0.813 & 0.779 & 0.797 & 0.847 \\
\rowcolor{green!45!white!10}
SAMed  & 0.862 & 0.710 &  0.798 & 0.677 & 0.735 & 0.944 & 0.766 & 0.874 & 0.798 & 0.775 & 0.579 & 0.790 & 0.776 \\
\rowcolor{green!45!white!10}
VMN  & 0.803 & 0.788 & 0.801 & 0.783 & 0.712 & 0.870 & 0.821 & 0.832 & 0.825 & 0.742 & 0.655 & 0.710 & 0.779 \\
\rowcolor{green!45!white!10}
FCFI  & 0.876 & 0.834 & 0.889 & 0.795 & 0.781 & 0.945 & 0.887 & 0.921 & 0.897 & 0.829 & 0.780 & 0.760 & 0.858 \\
\hline
% Med-SA & 0.972 & 0.945 & 0.958 & 0.843 & 0.834 & 0.960 & 0.898 & 0.900 & 0.851 & 0.824 & 0.775 & 0.811 & 0.881 \\ 
% SAM 2 (zero) & 0.366  & 0.724 & 0.642 & 0.772 & 0.501 & 0.270  & 0.415 & 0.040  & 0.179 & 0.630       & 0.682     & 0.746 & 0.516     \\
\rowcolor{cyan!40!white!20}
SAM 2  & 0.861 & 0.882 & 0.913 & 0.864 & 0.832 & 0.861 & 0.891 & 0.835 & 0.822 & 0.855 & 0.831 & 0.871 & 0.860 \\
\rowcolor{cyan!40!white!20}
TrackAny  & 0.818 & 0.762 & 0.760 & 0.805 & 0.730 & 0.824 & 0.841 & 0.829 & 0.815 & 0.780 & 0.701 & 0.728 & 0.783 \\
\rowcolor{cyan!40!white!20}
iMOS  & 0.801 & 0.738 & 0.759 & 0.844 & 0.762 & 0.855 & 0.861 & 0.843 & 0.828 & 0.810 & 0.744 & 0.672 & 0.793 \\
\rowcolor{cyan!40!white!20}
UniverSeg  & 0.824 & 0.862 & 0.889 & 0.774 & 0.835 & 0.918 & 0.826 & 0.899 & 0.831 & 0.804 & 0.819 & 0.818 & 0.842 \\
\rowcolor{cyan!40!white!20}
OnePrompt  & 0.881 & 0.869 & 0.894 & 0.900 & 0.868 & 0.891 & 0.908 & 0.837 & 0.829 & 0.850 & 0.832 & 0.870 & 0.869 \\
\hline
\rowcolor{blue!40!white!20}
\textbf{MedSAM-2} & 0.922 & 0.931 & 0.919 & 0.932 & 0.918 & 0.865 & 0.871 & 0.896 & 0.834 & 0.847 & 0.840 &  0.911 & \textbf{0.890} \\

\hline
\end{tabular}}\label{tab:btcv}
\vspace{-5pt}
\end{table*}

\begin{figure*}[t]
    \begin{center}
    %\framebox[4.0in]{$\;$}
\includegraphics[page=3,trim={0 0 7cm 8cm},clip,width=0.93\linewidth]{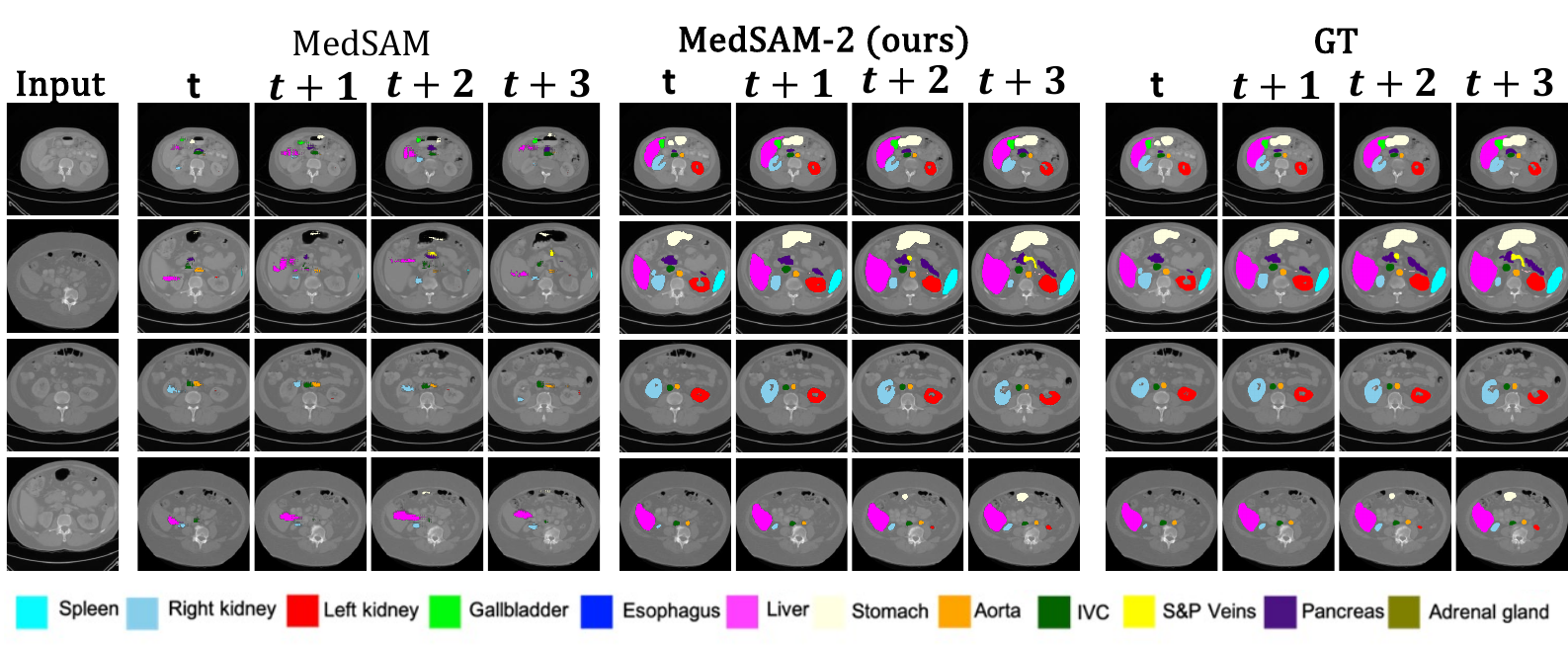}    \end{center}
\vspace{-15pt}
    \caption{\textbf{Qualitative Comparison on 3D Medical Image Segmentation.}  We show comparison of MedSAM \cite{MedSAM}, our MedSAM-2, and ground truth on sequential 3D medical image segmentation on the BTCV dataset \cite{fang2020multi}. Note how our MedSAM-2 produce more consistent 3D predictions leveraging the 3D context and maintaining high generalization capability compared to MedSAM \cite{MedSAM}.}\label{fig:vis}
\vspace{-10pt}
\end{figure*}

%%%%%%%%%%%%%%%%%%%%%%%%%%%%%%%%%

\begin{table*}[h]
\centering
% \vspace{-20pt}
\caption{\textbf{Universal 2D Medical Images Segmentation Performance.} We show the comparison of MedSAM-2 with task-tailored models, interactive generalized models, and auto-tracking generalized models. Evaluated on 11 unseen tasks by Dice Score (\%). }
\vspace{-6pt}
\resizebox{0.9\linewidth}{!}{%
\begin{tabular}{c|ccccccccccc|c}
\hline
Methods                                                      & KiTS         & ATLAS            & WBC          & SegRap           &  CrossM          & REFUGE          & Pendal          & LQN            & CAS          & CadVidSet            & ToothFairy                  & Ave            \\ \hline
TransUNet                                                    & 38.2          & 34.5          & 49.1          & 25.5          & 37.7          & 36.3          & 31.2          & 23.3          & 24.5          & 31.6          & 37.9                    & 33.6          \\ 
Swin-UNetr                                                  & 37.2          & 26.5          & 32.1          & 25.6          & 29.7          & 28.9          & 31.4          & 17.2          & 20.5          & 22.6          & 32.1                    & 28.5          \\ 
nnUNet                                                       & 39.8        & 30.3          & 40.4          & 26.8          & 35.0          & 34.9          & 42.9           & 18.9          & 37.4          & 41.8          & 35.3                  & 34.9          \\ 
MedSegDiff                                                   & 40.1          & 30.5          & 42.9          & 34.7          & 37.7          & 31.9          & 42.6          & 21.1          & 38.3          & 34.7          & 33.5                    & 35.3          \\  \hline
\begin{tabular}[c]{@{}c@{}} MSA\end{tabular}  & 54.6          & 48.9          & 55.9          & 47.3          & 51.7          & 49.2          & 54.2          & 41.0          & 48.9          & 53.5          & 47.6                    & 50.3          \\              
MedSAM  & 62.4          & 53.1          & 67.8          & 52.3          & 59.3          & 54.5          & 58.7          & 42.5          & 41.5          & 45.7          & 56.2                    & 53.9          \\ 
SAM-Med2D                                                       & 56.3          & 51.4          & 52.6          & 43.5          & 47.2          & 52.0          & 50.8          & 47.4          & 44.3          & 49.0          & 55.1                    & 50.0                  \\ \hline
SAM2  & 64.6          & 58.3          & 69.1          & 54.8          & 60.7          & 55.8          & 61.4          & 45.1          & 51.6          & 53.9          & 59.0                    & 57.6  \\
TrackAny   & 63.1          & 56.2          & 66.6          & 51.3          & 57.8          & 54.5          & 60.1          & 43.6          & 42.5          & 41.4          & 54.0                    & 53.7  \\
iMOS    & 62.6          & 53.8          & 61.4          & 52.5          & 54.3          & 57.0          & 58.7          & 42.2          & 44.8          & 46.1          & 50.7                    & 53.1 \\
UniverSeg   & 63.8  & 54.2 & 74.0  &  70.8  &  74.2  & 71.1          &  69.2  &  47.7  &  60.4  &  66.8  &  65.1   &  65.2 \\
One-Prompt  &  75.3  & 66.8 &  77.5  &  81.2  &  83.8  & 77.4          &  72.8  &  51.9  &  61.6  &  79.3  &  76.4   &  73.1  \\ 
%(&  72.6  &  49.5  &  64.5)

% \rowcolor{cyan!40!white!20} 
\rowcolor{blue!40!white!20}
MedSAM-2  &  \textbf{78.2}  & \textbf{71.8} &  \textbf{80.5}  &  \textbf{86.2}  &  \textbf{85.7}  & \textbf{79.9}  &  \textbf{76.0}  & \textbf{53.6}  & \textbf{66.5}  &  \textbf{86.1}  &  \textbf{80.8}   & \textbf{76.8}  \\ \hline
\end{tabular}%
}\label{tab:muti2d}
\vspace{-5pt}
\end{table*}

\begin{figure*}[t]
    \begin{center}
    %\framebox[4.0in]{$\;$}
% \includegraphics[page=4,trim={1cm 0 3cm 0},clip,width=0.9\linewidth]
\includegraphics[width=0.9\linewidth]{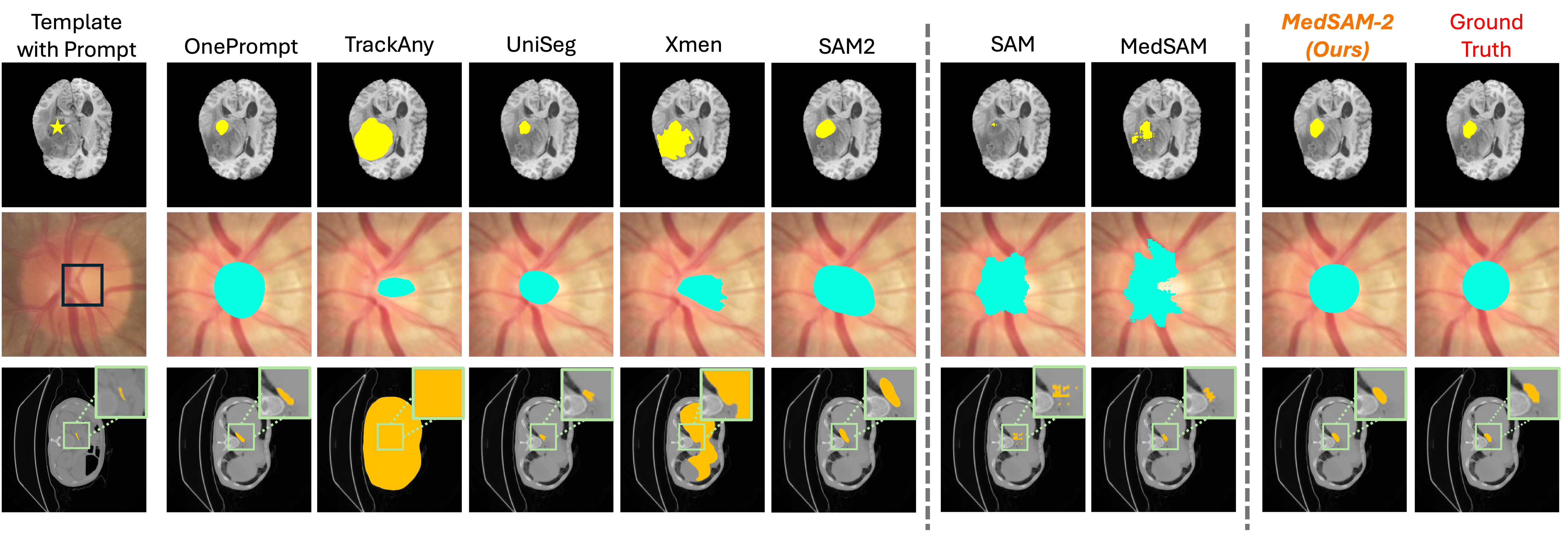}    \end{center}
\vspace{-15pt}
    \caption{\textbf{Qualitative Examples of MedSAM-2 for 2D One-Prompt Segmentation \& 3D Segmentation.} We show several examples of 2D segmentation on diverse datasets. }\label{fig:examples}
\end{figure*}

\begin{figure*}[t]
    \centering
    \includegraphics[width=\linewidth]{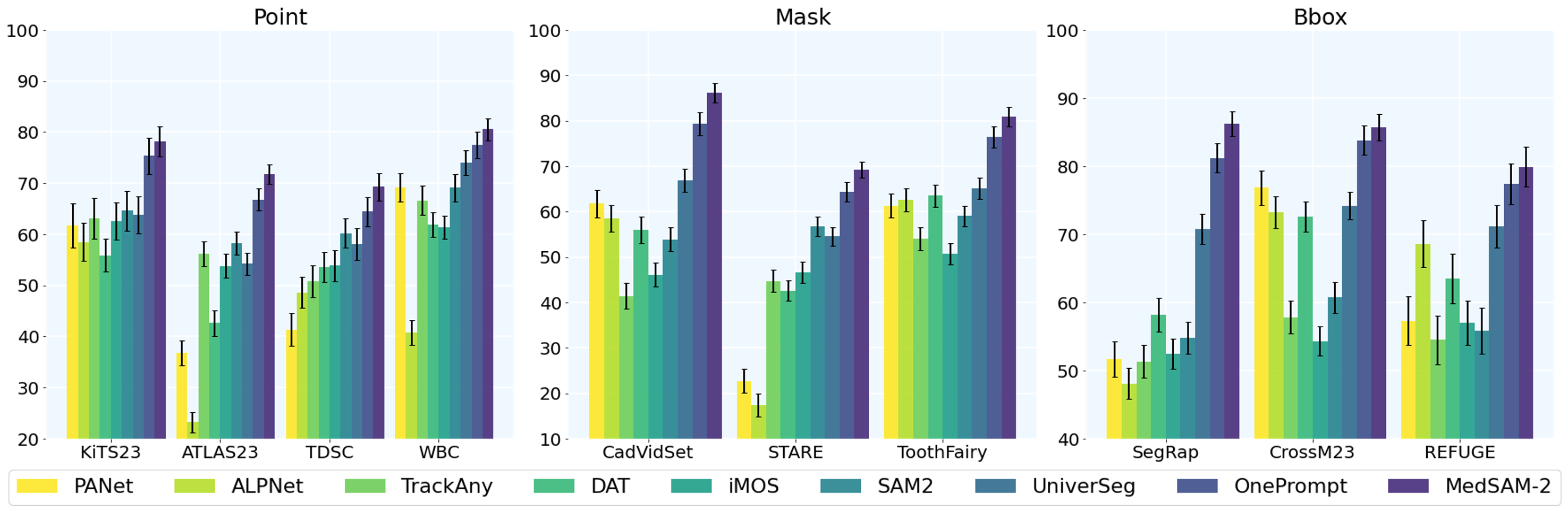}
    \vspace{-15pt}
    \caption{\textbf{One-prompt 2D Segmentation Performance.} We show MedSAM-2 \textit{v.s.} Few/One-shot Models under One-prompt Segmentation setting on 10 datasets with different prompts. Our MedSAM-2 colored by the darkest blue on the right of each bar group. }
    \label{fig:trans_pkg}
    \vspace{-12pt}
\end{figure*}

\section{Experiment}
\label{headings}
\subsection{Dataset}
To build a foundation model with strong generalization on unseen tasks, we train and test our model on the One-Prompt dataset \cite{Wu_2024_CVPR}, a large-scale and diverse collection of 2D and 3D medical images assembled from publicly accessible datasets with clinicians-annotated prompts. This data source comprises 78 datasets across various medical domains and imaging modalities, covering a wide range of organs such as lungs \cite{saporta2021deep, setio2017validation, simpson2019large}, eyes \cite{fang2022refuge2, stare, ma2020rose, orlando2020refuge}, brain \cite{baid2021rsna, gollub2013mcic, hernandez2022isles, kuijf2019standardized, kuklisova2011dynamic}, and abdominal organs \cite{ji2022amos, bloch2015nci, heller2021state, kavur2021chaos, lambert2020segthor, landman2015miccai, lemaitre2015computer, litjens2014evaluation, luo2021word, ma2021abdomenct, radau2009evaluation, simpson2019large}. Each dataset includes at least one image or slice annotated by a professional clinician, with over 3,000 samples collectively annotated by clinicians. A detailed list of the One-Prompt datasets is provided in the supplementary materials.

We follow the default split of the One-Prompt dataset, using 64 datasets for training and 14 for testing. The test set includes 8 MICCAI2023 Challenge tasks across diverse anatomies—kidney tumor \cite{heller2023kits21}, liver tumor \cite{quinton2023tumour}, breast cancer \cite{tdsc-abus2023}, nasopharynx cancer \cite{astaraki2023fully}, vestibular schwannoma \cite{CrossMoDA23}, mediastinal lymph node \cite{LNQ2023}, cerebral artery \cite{CAS2023}, and inferior alveolar nerve \cite{ToothFairy}—along with 6 other tasks for structures like white blood cells \cite{WBC}, optic cups \cite{fang2022refuge2}, mandibles \cite{pendal}, coronary arteries \cite{CadVidSet}, abdominal organs \cite{fang2020multi}, and retinal vessels \cite{stare}. We evaluate model performance on each test dataset using task-specific prompts: the \textit{Click} prompt for KiTS23, ATLAS23, TDSC, and WBC; the \textit{BBox} prompt for SegRap, CrossM23, REFUGE, Pendal, LNQ23, and CAS23; and the \textit{Mask} prompt for CadVidSet, STAR, BTCV-test and ToothFairy. Among these, STAR, BTCV-test, and TDSC involve tasks seen in training, while the remaining 11 tasks are used for zero-shot testing.

\subsection{Human-User Prompted Evaluation}
For evaluation, we engaged human users to simulate real-world interactions in prompt-based segmentation. Fifteen users were assigned to prompt approximately 10\% of the test images, including 5 laypersons with a clear understanding of the task but no clinical background, 7 junior clinicians, and 3 senior clinicians. This setup aims to reflect real-world prompting scenarios, such as clinical training or semi-automated annotation.

\subsection{Implementation}
We conduct training and testing on the PyTorch platform, leveraging 64 NVIDIA A100 GPUs for distributed training. Optimization uses the AdamW optimizer ($\beta_{1} = 0.9$, $\beta_{2} = 0.999$) with a linear learning rate warmup followed by cosine decay. Our training simulates an interactive environment by sampling 8-frame sequences, randomly selecting up to 2 frames (including the first) for corrective clicks. Prompts are generated from ground-truth masks and model predictions, with initial prompts consisting of the ground-truth mask with 50\% probability, a positive click from the mask with 25\%, or a bounding box input with 25\%. To maintain diversity across tasks and prompts, we use a balanced sampling strategy that avoids equal representation across all tasks, as certain image modalities, tasks, or prompt types are more frequent. To prevent overfitting to these dominant elements, we uniformly select tasks and sequence states, starting with a random task selection, then narrowing the pool to data associated with that task. We proceed by selecting an image modality available for the task, refining the pool to ensure homogeneity, and finally selecting a sample from the filtered set. 
All comparison models are trained and tested under the same setting. Additional details on data processing and training are provided in the supplementary material. 

% \subsection{Evaluation Metrics}
% \mysection{Intersection over Union (IoU)}
% also known as the Jaccard Index, is a measure used to evaluate the accuracy of an object detector on a dataset. It quantifies overlap by dividing the area of overlap between the predicted segmentation and the ground truth by the area of their union \cite{swin-unetr}. 

% \mysection{Dice Score}
% or Dice Coefficient is a statistical tool that compares the similarity between two samples. It is particularly prevalent in medical image analysis due to its sensitivity to the size of the objects being examined \cite{MedSAM}. The Dice Score is calculated by taking twice the area of overlap between the predicted and actual segmentations, divided by the total number of pixels in both the prediction and the ground truth.

% \mysection{Hausdorff Distance (HD95) Metric}
% The Hausdorff Distance (HD95) is a metric to determine the extent of discrepancy between two sets of points, typically used to evaluate the accuracy of object boundaries in image segmentation tasks. It is particularly useful for quantifying the worst-case scenario of the distance between the predicted segmentation and the ground truth boundary \cite{karimi2019reducing}.
% While the Hausdorff Distance provides a strict measure by considering the maximum distance, it can be overly sensitive to outliers. To mitigate this, the HD95 metric considers only the 95th percentile of the distances instead of the maximum \cite{Wu_2024_CVPR}. 

%% file: sec/5.results.tex
\section{Results} \label{sec:results}
In this section, we present a comprehensive evaluation of MedSAM-2 on both 2D and 3D medical image segmentation tasks. We compare our model with a range of state-of-the-art (SOTA) methods, including task-specific, diffusion-based, and interactive segmentation models. Performance is quantified using the Dice coefficient, Intersection over Union (IoU), and Hausdorff Distance (HD95) where appropriate.
% Note that the comparison presented may not be entirely fair due to undisclosed training details, such as data and implementation, for some models. Additionally, differences in model functionalities and intended application scenarios mean that many models cannot be compared on equal footing. Therefore, we treat these models primarily as references and focus more on presenting our approach for building a powerful medical image segmentation model.

\subsection{Performance of Universal Medical Image Segmentation}
% In this section, we benchmark MedSAM-2 against a comprehensive array of state-of-the-art (SOTA) medical image segmentation methods, covering both 2D and 3D medical image segmentation tasks widely recognized within the community.
For 3D medical images, prompts are provided to frames with a probability of 0.25, meaning each frame has a 25\% likelihood of receiving a prompt. For 2D images, prompts are provided with a probability of 0.3. Results for 3D medical image segmentation are presented in Table \ref{tab:btcv}, while 2D results are presented in Table \ref{tab:muti2d}.

\mysection{On 3D Medical Images} To assess the general performance of MedSAM-2 on 3D medical images, we conducted experiments on the BTCV multi-organ segmentation dataset (\figLabel{\ref{fig:vis}}). We compare MedSAM-2 with established SOTA segmentation methods such as nnUNet~\cite{isensee2021nnu}, TransUNet~\cite{chen2021transunet}, UNetr~\cite{hatamizadeh2022unetr}, Swin-UNetr~\cite{hatamizadeh2022swin}, and diffusion-based models like EnsDiff~\cite{wolleb2021diffusion}, SegDiff~\cite{amit2021segdiff}, and MedSegDiff~\cite{wu2022medsegdiff}. Additionally, we evaluate interactive segmentation models including SAM~\cite{kirillov2023segment}, MedSAM~\cite{MedSAM}, SAMed~\cite{samed}, SAM-Med2D~\cite{sam-med2d}, SAM-U~\cite{sam-u}, VMN~\cite{VMN}, and FCFI~\cite{fcfi}. For FCFI, ConvNext-v2-H~\cite{woo2023convnext} is used as the backbone. We also compare MedSAM-2 with auto-tracking generalized models, such as SAM 2 \cite{ravi2024sam2}, TrackAny \cite{yang_track_2023}, iMOS \cite{yan_foundation_2024}, UniverSeg \cite{butoi2023universeg}, OnePrompt \cite{Wu_2024_CVPR}. Table~\ref{tab:btcv} presents the quantitative results on the BTCV dataset. MedSAM-2 achieves a Dice score of 89.0\%, outperforming all compared methods. Specifically, MedSAM-2 surpasses the previous SOTA model MedSegDiff by a margin of 1.10\%. Among interactive models, MedSAM-2 maintains the lead, outperforming the previously leading interactive model, FCFI, by 3.20\%. It is important to note that all these competing interactive models require prompts for each frame, whereas MedSAM-2 achieves better results with far fewer user prompts.

\begin{figure}[t]
    \begin{center}
    %\framebox[4.0in]{$\;$}
% \includegraphics[trim={0 10.2cm 40cm 0},clip, width=0.8\linewidth]{medsam2/fig/memory_bank.png} 
\includegraphics[trim={0 0 0 0},clip, width=1.0\linewidth]{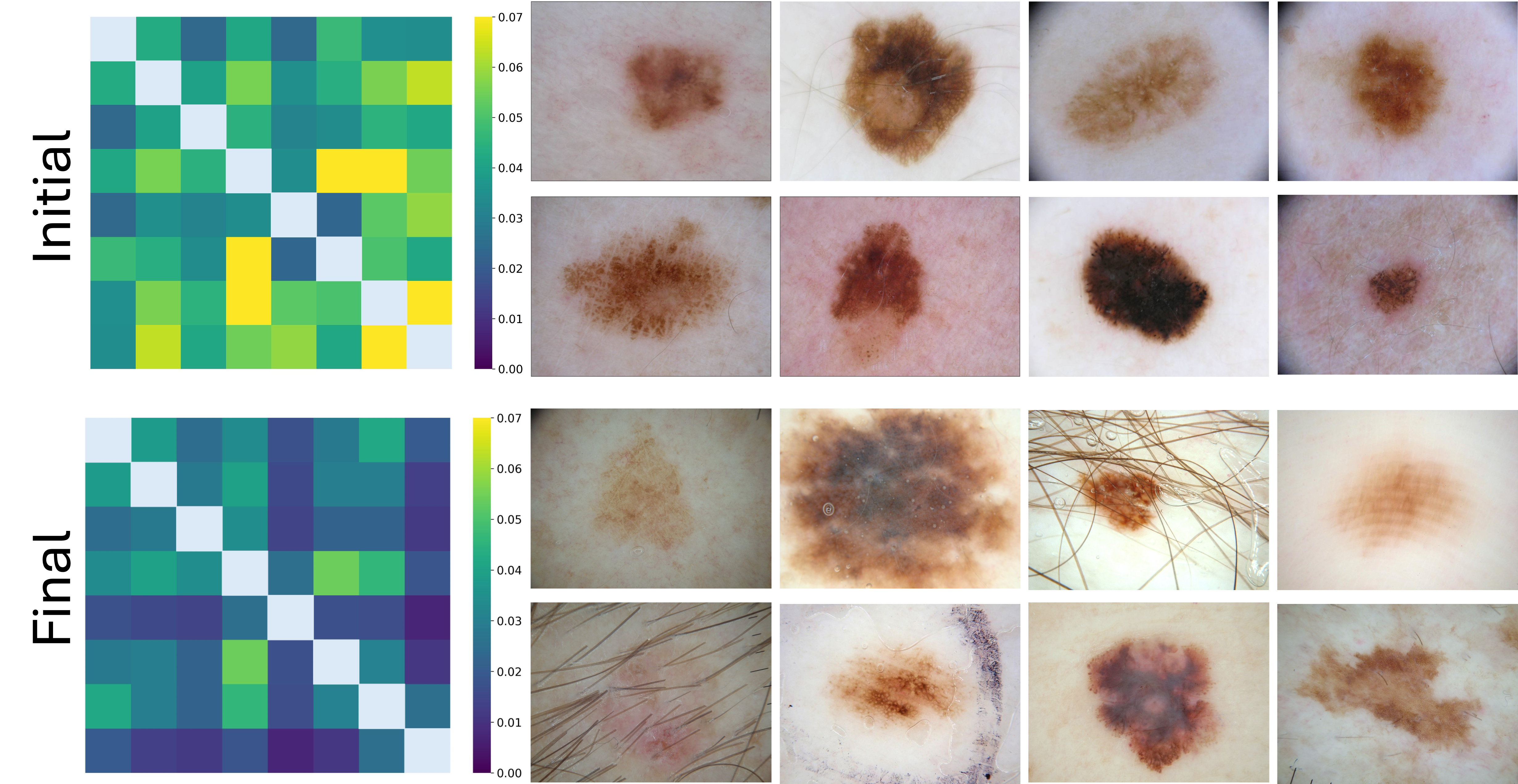}
\end{center}
\vspace{-8pt}
    \caption{\textbf{Mutual Information Analysis of the Self-Sorting Memory Bank.} We show pairwise mutual information analysis and visualizations of memory bank samples at the start and final stages. The total mutual information between the memory bank samples decreases from 2.54 in the initial stage to 1.43 in the final stage. }\label{fig:memory_bank}
    \vspace{-18pt}
\end{figure}

\mysection{On 2D Medical Images} 
We further evaluate MedSAM-2 in a zero-shot setting on 11 unseen 2D medical image segmentation tasks.
%, including optic disc/cup segmentation (REFUGE2 dataset), brain tumor segmentation (BraTS dataset), thyroid nodule segmentation (TNMIX dataset), and skin lesion segmentation (ISIC dataset). We compare MedSAM-2 with task-specific models such as ResUNet~\cite{yu2019robust}, BEAL~\cite{wang2019boundary}, TransBTS~\cite{wang2021transbts}, SwinBTS~\cite{wang2021transbts}, MTSeg~\cite{gong2021multi}, UltraUNet~\cite{chu2021ultrasonic}, FAT-Net~\cite{wu2022fat}, and BAT~\cite{wang2021boundary}. Additionally, 
Similar to 3D medical image segmentation ask, we compare the results with task-tailored models, interactive models that require prompts for each image, and auto-tracking models.
% We also evaluate MedSAM-2 against state-of-the-art (SOTA) segmentation methods tailored for specific tasks across different image modalities, as detailed in Table~\ref{tab:muti}. For optic cup segmentation, we compare with ResUnet\cite{yu2019robust} and BEAL\cite{wang2019boundary}; for brain tumor segmentation, with TransBTS\cite{wang2021transbts} and SwinBTS\cite{wang2021transbts}; for thyroid nodule segmentation, with MTSeg\cite{gong2021multi} and UltraUNet\cite{chu2021ultrasonic}; and for skin lesion segmentation, with FAT-Net\cite{wu2022fat} and BAT\cite{wang2021boundary}. Additionally, we benchmark against interactive models that require user prompts for each image.
Table~\ref{tab:muti2d} summarizes the results across different datasets. MedSAM-2 consistently outperforms the compared methods,  demonstrating its superior generalization capability across diverse medical imaging modalities. For instance, MedSAM-2 improves the Dice score by 2.5\% on optic disc segmentation and 2.9\% on brain tumor segmentation compared to the previous best models. Even when compared to interactive models that require prompts for each image, MedSAM-2 maintains its lead, highlighting the effectiveness of our proposed self-sorting memory bank mechanism. See \figLabel{\ref{fig:examples}} for visualizations.

% The results from the table illustrate that MedSAM-2 outperforms all compared methods across five distinct tasks, showcasing its superior generalization across diverse medical segmentation tasks and image modalities. Specifically, MedSAM-2 achieves improvements of 2.0\% on the Optic-Cup, 1.6\% on Brain Tumor, and 2.8\% on Thyroid Nodule, measured by the Dice score. Even against interactive models—which necessitate prompts for each image—MedSAM-2 maintains its lead, affirming the effectiveness of the proposed confidence memory bank in boosting performance.

\subsection{One-prompt Segmentation Performance under different prompts}
We further assess MedSAM-2 under the One-Prompt segmentation setting by comparing it to various few/one-shot learning baselines that use different prompts. 
%This experiment evaluates MedSAM-2's performance in a challenging scenario where there is no clear connection between sequential images.
We benchmark against few/one-shot models such as PANet~\cite{wang2019panet}, ALPNet~\cite{ouyang2020self},TrackAny~\cite{yang_track_2023}, DAT~\cite{zhao2019data}, iMOS~\cite{yan_foundation_2024}, turned SAM2~\cite{ravi2024sam2}, UniverSeg~\cite{butoi2023universeg}, and One-Prompt~\cite{Wu_2024_CVPR}. We further evaluate the models by testing them 5 times with different prompted images and input sequences to observe performance variance. Figure~\ref{fig:trans_pkg} presents the average Dice scores and variance per task for each method. MedSAM-2 not only consistently achieves higher average performance but also demonstrates significantly lower variance in most cases, underscoring its robust generalization across various tasks and prompt types.

\begin{figure}[t]
    \begin{center}
\includegraphics[width=1\linewidth]{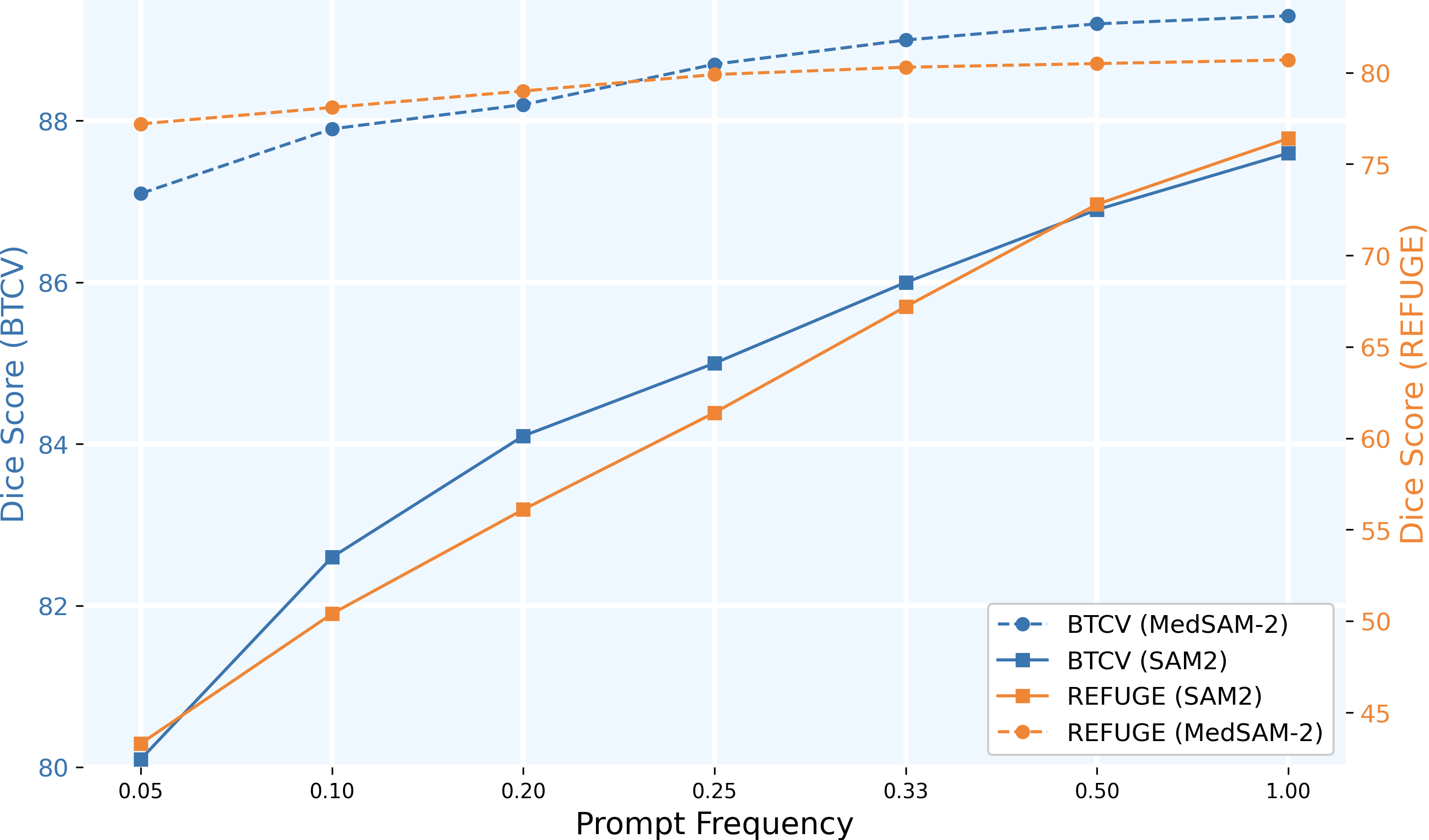}    \end{center}
\vspace{-8pt}
    \caption{\textbf{Impact of Prompt Frequency on 2D and 3D Medical Images.} We report the Dice Score (\%) of  MedSAM-2 on REFUGE (2D) and BTCV (3D) datasets with varying prompt frequencies.}\label{fig:prompt_freq}
\vspace{-10pt}
\end{figure}

\subsection{Analysis and Ablation Study}
\mysection{Mutual Information Analysis of Memory Bank}
We analyze the effectiveness of the self-sorting memory bank in MedSAM-2 by examining the mutual information of stored embeddings over time using the ISIC dataset. The mutual information analysis assesses the diversity of embeddings in the memory bank.
% We evaluate the memory bank's diversification process using the ISIC dataset by examining pairwise mutual information across different stages. This analysis assesses the memory bank’s ability to capture a progressively diverse set of features, transitioning from relatively similar samples to a broad range of distinct representations over time. 
Figure~\ref{fig:memory_bank} illustrates the pairwise mutual information of memory bank samples at different stages. Initially, the total mutual information is high (2.54), indicating redundancy among the stored embeddings. As the memory bank evolves, the mutual information decreases to 1.43, showing that the embeddings become more diverse and representative of different features. This confirms that the self-sorting mechanism effectively captures a diverse set of informative embeddings, enhancing the model's generalization capability. The right side of Figure \ref{fig:memory_bank} visually supports this trend, showing an increasingly diverse set of samples as the memory bank develops from the initial to the final stage.

\mysection{Prompt Frequency Analysis on 2D and 3D Medical Images}
We conduct experiments to study the impact of prompt frequency on the performance of 2D (REFUGE) and 3D (BTCV) datasets. The performance improves with increasing prompt given frequency, as seen in the progressive increase in Dice scores for both datasets (Figure \ref{fig:prompt_freq}). Compared to SAM 2, our model demonstrates greater robustness under varying prompt frequencies. On 3D images, the performance gap between 5\% prompting and full prompting is only 2\% for our model, while SAM 2 shows a 7.5\% gap. This difference is even more pronounced in 2D medical images, where our model maintains a 3.5\% gap, whereas SAM 2 shows a substantial 33.1\% drop. This highlights how our self-sorting memory bank significantly enhances model robustness, achieving strong performance even with minimal human interaction.

% The left of Figure \ref{fig:memory_bank} illustrates the pairwise mutual information of memory bank samples, showing a clear decrease from intinal stage to final stage of the . Initially, the total mutual information is 2.54, indicating substantial redundancy, as the stored samples are relatively similar with limited feature diversity. As the memory bank evolves, this mutual information reduces significantly to 1.43 in the final stage, demonstrating the effectiveness of our approach in capturing a more diverse and distinct set of samples. This consistent reduction in mutual information confirms that the memory bank becomes progressively more varied, with samples growing less similar over time. The right side of Figure \ref{fig:memory_bank} visually support this trend, showing an increasingly diverse set of samples as the memory bank develops from the initial to the final stage.

\mysection{Ablation Study}
In the ablation study, we evaluate several key design choices for the MedSAM-2 model, including the use of an IOU confidence threshold $c_{\text{thresh}}$ from Section {\ref{sec:medsam2}} for storing samples, the storage of dissimilar templates in the memory bank, and the application of resampling strategies on the memory bank. This study is conducted with with CadVidSet dataset and the aorta task in the BTCV dataset.
Table~\ref{tab:ablation} presents the results of the ablation study. Using an IOU confidence threshold for the memory bank improves CadVidSet dataset's average Dice score to 57.8\%, whereas without the threshold, it reaches only 53.9\%. This selective storage based on confidence enhances the quality of retained samples and reduces redundancy. In addition, further storing dissimilarity templates results in a Dice score of 64.5\% and 88.4\% for CadVidSet and BTCV dataset, respectively. The dissimilarity-based storage captures a broader range of features, allowing the memory bank to adapt more effectively to diverse inputs. Applying the resampling conditioned on feature relevance raises the Dice score to 72.9\% and 89.6\% for CadVidSet and BTCV datasets, by prioritizing relevant samples for specific segmentation tasks. 
% These results underscore the importance of adaptive storage and retrieval strategies in building a diverse and effective memory bank, with each design choice contributing to enhanced segmentation performance. 
 %We study further design choices like the choice of the 6 directions of treating 3D in \secLabel{\ref{sec:unified}} and the choice of the IOU threshold $c_{\text{thresh}}$ from \secLabel{\ref{sec:medsam2}} in \supp.

% {\color{red}2. add detailed ablation in supplementary for 3D case, 1-6 directions}

% {\color{red}3. add detailed ablation in supplementary for IOU threshold chosen}

%------------------------------------------------------------------------
\begin{table}[t]
\centering
% \vspace{-10pt}
\caption{\textbf{Ablation Study of MedSAM-2.} We evaluate each of the novel components of MedSAM-2 pipeline on the CadVidSet dataset \cite{CadVidSet} and the aorta task in BTCV dataset \cite{fang2020multi}.}
% \vspace{5pt}
\resizebox{1\linewidth}{!}{%
\begin{tabular}{ccccc}
\hline\hline
IOU Confidence & Dissimilar & Memory & \multirow{2}{*}{CadVidSet} & \multirow{2}{*}{BTCV-Aorta} \\
Threshold &  Templates & Resampling & & \\ \hline
 &   &   &   53.9 & 83.5  \\
\checkmark &   &   &   57.8 & 86.2  \\
\checkmark & \checkmark  &          &  64.5 & 88.4  \\
% \checkmark &           & \checkmark &  ? & ?\\
%  &    \checkmark       & \checkmark &  72.9 & 89.6\\
% \rowcolor{cyan!40!white!20}
\rowcolor{blue!40!white!20}
\checkmark &    \checkmark   & \checkmark & \textbf{72.9} &  \textbf{89.6} \\ \hline \hline
\end{tabular}%
}\label{tab:ablation}
\vspace{-10pt}
\end{table}
%-------------------------------------------------------------------------

%% file: sec/6_conclusion.tex
\section{Conclusion}

In this work, we introduced MedSAM-2, an generalized auto-tracking segmentation model for both 2D and 3D medical images. By treating medical images as videos and incorporating a novel \emph{self-sorting memory bank}, MedSAM-2 effectively handles unordered medical images and enhances generalization across diverse tasks. 
%The self-sorting memory bank dynamically selects informative embeddings based on confidence and dissimilarity, enabling the model to adapt to new targets with minimal user interaction. 
This innovation unlocks the \emph{One-Prompt Segmentation} capability, allowing MedSAM-2 to generalize from a single prompt to segment similar structures across multiple images without temporal relationships. Comprehensive evaluations across 14 benchmarks and 25 tasks demonstrate that MedSAM-2 consistently outperforms state-of-the-art models in both 2D and 3D medical image segmentation. It achieves superior performance while reducing the need for continuous user interaction, making it particularly advantageous in clinical settings.

%% file: sec/X_suppl.tex
\clearpage
\setcounter{page}{1}
% \maketitlesupplementary

% \label{sec:rationale}
% % 
% Having the supplementary compiled together with the main paper means that:
% % 
% \begin{itemize}
% \item The supplementary can back-reference sections of the main paper, for example, we can refer to \cref{sec:intro};
% \item The main paper can forward reference sub-sections within the supplementary explicitly (e.g. referring to a particular experiment); 
% \item When submitted to arXiv, the supplementary will already included at the end of the paper.
% \end{itemize}
% % 
% To split the supplementary pages from the main paper, you can use \href{https://support.apple.com/en-ca/guide/preview/prvw11793/mac#:~:text=Delete%20a%20page%20from%20a,or%20choose%20Edit%20%3E%20Delete).}{Preview (on macOS)}, \href{https://www.adobe.com/acrobat/how-to/delete-pages-from-pdf.html#:~:text=Choose%20%E2%80%9CTools%E2%80%9D%20%3E%20%E2%80%9COrganize,or%20pages%20from%20the%20file.}{Adobe Acrobat} (on all OSs), as well as \href{https://superuser.com/questions/517986/is-it-possible-to-delete-some-pages-of-a-pdf-document}{command line tools}.

\section{Why Does Self-Sorting Work?} \label{secsup:self-sorting}

The effectiveness of the self-sorting memory bank $\mathcal{M}^{\text{sort}}$ in MedSAM-2 can be understood through the lens of information theory, particularly in terms of entropy and mutual information.

Let $x_t$ denote the input image at time $t$, let $Y_t$ denote the predicted segmentation mask at time $t$, and let $Z_t$ denote the ground truth at time $t$. The mutual information $I(Y_t;Z_t|x_t)$ measures the amount of information that the predicted segmentation mask contain about the ground truth:
\begin{equation}
    I(Y_t; Z_t|x_t) = I(\mathcal{D}(\mathcal{A}(\mathbf{F}_t(x_t),\mathcal{M}_t,\mathbf{Q}_t)); Z_t|x_t)
\end{equation}
Given that the input image $x_t$ is specified, both $\mathbf{F}_t$ and $\mathbf{Q}_t$ remain constant. Consequently, the only variable is the selected memory bank $\mathcal{M}_t$. Therefore, increasing the mutual information between $\mathcal{M}_t$ and $Y_t$, conditioned on $x_t$, will lead to an improved predicted mask.

By leveraging the relationship between mutual information and conditional entropy, the following decomposition can be derived:
\begin{equation}
    I(M_t;Z_t|x_t)=H(Z_t)-H(Z_t|M_t,x_t)
\end{equation}
where $H(Z_t|x_t)$ denotes the entropy of the ground truth given the input image, and $H(Z_t|M_t,x_t)$ represents the conditional entropy of the ground truth $Z_t$ given the known $M_t$, defined as:
\begin{align}
    H(Z_t|M_t,x_t)&=-\sum_{x\in\mathcal{X},y\in\mathcal{Y}}p(x,y)\text{log}\frac{p(x,y)}{p(x)}\\
    &=\mathbb{E}[-\text{log}\frac{p(x,y)}{p(x)}]\\
    &=\mathbb{E}[-\text{log}(p(y|x))]
\end{align}
Since $-\text{log}(p(y|x))$ can be interpreted as the amount of information required to describe the random variable $y$ given the value of $x$, the conditional entropy $H(Z_t|M_t,x_t)$ can thus be viewed as the expected information needed to describe the ground truth $Z_t$ given the selected memory bank $M_t$.

By selecting embeddings based on the highest confidence scores, the self-sorting memory bank seeks to maximize $I(M_t; Z_t|x_t)$ by minimizing $H(Z_t | M_t,x_t)$. Given that $H(Z_t|x_t)$ is assumed to be constant, high-confidence embeddings provide more information regarding the output, thereby reducing the information required to describe $Z_t$. This reduction leads to a smaller $H(Z_t | M_t,x_t)$ and, consequently, an increase in mutual information. This increase in mutual information suggests that the model is able to make more accurate and reliable predictions based on the embeddings stored in the memory bank.

Furthermore, the self-sorting mechanism introduces variability in the selection of memory embeddings, as it is not limited to the most recent frames but instead selects from all past frames based on confidence scores. This variability enhances the diversity of information within the memory bank $M_t$, potentially decreasing the additional information needed to infer $Z_t$, especially in contexts where frames change rapidly and significantly. As a result, the self-sorting mechanism can yield a smaller $H(Z_t | M_t,x_t)$, thereby increasing the mutual information $I(M_t; Z_t|x_t)$.

This increased entropy in the memory embeddings enhances the model's ability to generalize. According to the principle of maximum entropy~\cite{jaynes1957information}, a model that considers a broader distribution of features is less likely to overfit to specific patterns in the training data and is better equipped to handle variability in unseen data. By increasing both the mutual information between the memory embeddings and the output and the entropy of the memory embeddings themselves, the self-sorting memory bank improves the robustness and generalization of MedSAM-2.

Consequently, the model is better suited to handle unordered medical images, as it leverages the most informative and diverse embeddings for segmentation. This leads to enhanced performance across diverse medical imaging tasks after training with standard segmentation loss~\cite{MedSAM}.

%%%%%%%%%%%%%%%%%%%%%%%%%%%%%%%%%%%%%%%%%%%%%%%%%%%%%%%%%%%%%%%%%%%%%%%%%%5
\section{Experimental Details} \label{secsup:expdetails}
\subsection{Evaluation Metrics}
We use Intersection over Union (IoU) and Dice Score to assess the performance of models in medical image segmentation.

\paragraph{Intersection over Union (IoU)}
Intersection over Union (IoU), also known as the Jaccard Index, is a measure used to evaluate the accuracy of an object detector on a specific dataset. It quantifies the overlap between two datasets by dividing the area of overlap between the predicted segmentation and the ground truth by the area of their union. The formula for IoU is given by:
\[
\text{IoU} = \frac{\text{Area of Overlap}}{\text{Area of Union}}
\]
IoU provides a clear metric at the object level, assessing both the size and position accuracy of the prediction relative to the actual data, which is particularly useful for understanding detection model performance.

\paragraph{Dice Score}
The Dice Score, or Dice Coefficient, is a statistical tool that compares the similarity between two samples. It is particularly prevalent in medical image analysis due to its sensitivity to the size of the objects being examined. The Dice Score is calculated by taking twice the area of overlap between the predicted and actual segmentations, divided by the total number of pixels in both the prediction and the ground truth. The formula for the Dice Score is:
\[
\text{Dice Score} = \frac{2 \times \text{Area of Overlap}}{\text{Area of Prediction} + \text{Area of Ground Truth}}
\]
This score ranges from 0 to 1, where a score of 1 indicates perfect agreement between the model's predictions and the ground truth. The Dice Score is known for its robustness against the size variability of the segmented objects, making it extremely valuable in medical applications where such variability is common.

Both metrics, IoU and Dice Score, provide comprehensive insights into model accuracy, with Dice Score being particularly effective in scenarios involving significant variations in object size.

\paragraph{Hausdorff Distance (HD95) Metric}

The Hausdorff Distance (HD95) is a metric used to determine the extent of discrepancy between two sets of points, typically used to evaluate the accuracy of object boundaries in image segmentation tasks. It is particularly useful for quantifying the worst-case scenario of the distance between the predicted segmentation and the ground truth boundary.

The Hausdorff Distance measures the maximum distance of a set to the nearest point in the other set. For image segmentation, this means calculating the greatest of all the distances from a point in the predicted boundary to the closest point in the ground truth boundary and vice versa. The formula for the Hausdorff Distance is given by:
\[
\text{HD} = \max\left(\sup_{x \in A} \inf_{y \in B} d(x,y), \sup_{y \in B} \inf_{x \in A} d(x,y)\right)
\]
where \(A\) and \(B\) represent the sets of boundary points of the ground truth and the predicted segmentation, respectively, and \(d(x,y)\) denotes the Euclidean distance between points \(x\) and \(y\).

While the Hausdorff Distance provides a strict measure by considering the maximum distance, it can be overly sensitive to outliers. To mitigate this, the HD95 metric is used, which considers only the 95th percentile of the distances instead of the maximum. This adjustment makes the HD95 less sensitive to outliers and provides a more robust measure for practical applications:
\[
\text{HD95} = \text{95th percentile of } \{d(x,y) \mid x \in A, y \in B\}
\]
This metric is particularly relevant in medical image analysis where precision in the segmentation of anatomical structures is critical and outliers can distort the evaluation of segmentation performance.

% section a: dataset
\section{Data}
% \subsection{Data Source}
% We provide the details of our data source in Table \ref{tab:source}.

\subsection{Data Preprocessing}
The original 3D datasets contain a variety of CT and MRI images stored in DICOM, NRRD, or MHD formats. To ensure uniformity and compatibility, all images, regardless of modality, were converted to the widely used NIfTI format. This conversion also included grayscale images, such as X-Ray and Ultrasound, while RGB images depicting endoscopy, dermoscopy, fundus, and pathology were converted into the PNG format. For tasks involving multiple segmentation targets, each target is treated as an individual task for predicting a binary segmentation mask. During the inference stage for predicting multiple targets, we predict a soft segmentation mask with a fixed threshold (averaging 0.5) to filter out uncertain predictions.

Notably, image intensities varied significantly across modalities. For instance, CT images ranged from $-2000$ to $2000$, MRI values ranged from $0$ to $800$, endoscopy/ultrasound images from $0$ to $255$, and some modalities were already within the range $0$ to $1$. To harmonize this variability, intensity normalization was systematically conducted for each modality. The default normalization during training and inference involved normalizing each image independently by subtracting its mean and dividing by its standard deviation. For MRI, X-Ray, ultrasound, mammography, and Optical Coherence Tomography (OCT) images, intensity values were trimmed to fall between the 0.5th and 99.5th percentiles before normalization. If cropping resulted in a $25\%$ or greater reduction in average size, a mask for central non-zero voxels was generated, and normalization was confined to this mask, disregarding surrounding zero voxels. For CT images, Hounsfield units were first normalized using window width and level values before applying standard normalization. Furthermore, since CT intensity values quantitatively reflect tissue properties, we applied a global normalization scheme to all images. Specifically, this involved clipping intensity values to the 0.5th and 99.5th percentiles of foreground voxels, followed by normalization using the global foreground mean and standard deviation.

To standardize image sizes, the provided samples were first cropped to their non-zero regions and then uniformly resized to $256 \times 256$. During resizing, we used bi-cubic interpolation for images and nearest-neighbor interpolation for masks, ensuring smooth standardization and compatibility across all images. For 3D images, we generally operated on the two axes with the highest resolution. If all three axes were isotropic, the two trailing axes were used for slice extraction. The channel was replicated threefold to ensure consistency during processing. For slice-based processing, no resampling along the out-of-plane axis was required.

Masks with multiple classes were processed into individual masks for each class. Masks containing multiple connected components were dissected, while original masks were retained in cases with only one component. Additionally, masks where the target area was less than $0.153\%$ of the total image (equivalent to areas smaller than $100$ pixels in a resized $256 \times 256$ resolution) were excluded. This deliberate decision ensures the dataset only includes significant and well-defined target areas. The standardized preprocessing pipeline was consistently applied across all compared methods to ensure a fair and unbiased comparison.

\subsection{Data Augmentation}
During training, we utilize a range of data augmentation techniques, dynamically computed on the CPU. Spatial augmentations are applied, including rotations, scaling, Gaussian noise, Gaussian blur, intensity and contrast adjustments, low-resolution simulation, gamma correction, and flipping. To enhance image variability, most augmentations involve random parameter selection from predefined ranges. The application of these augmentations follows stochastic principles, adhering to predefined probabilities. Consistent augmentation parameters are maintained across datasets. Each augmentation is individually applied to both the template sample and the query sample.

Details of the augmentation techniques are as follows:

\begin{enumerate}
  \item \textbf{Rotation:} Applied with a probability of 0.15 to all images. The rotation angle is uniformly sampled from the range \([-25, 25]\).

  \item \textbf{Scaling:} Scaling is achieved by multiplying image coordinates with a scaling factor. Scale factors smaller than 1 result in a "zoom out" effect, while values larger than 1 create a "zoom in" effect. The scaling factor is uniformly sampled from \([0.7, 1.4]\), with a probability of 0.15.

  \item \textbf{Gaussian Noise:} Zero-centered Gaussian noise is independently added to each sample with a probability of 0.15. The noise variance is sampled from \([0, 0.1]\), considering that normalized sample intensities are close to zero mean and unit variance.

  \item \textbf{Gaussian Blur:} Blurring is applied with a probability of 0.15 per sample. For each task, it occurs with a probability of 0.5 per modality. The Gaussian kernel size is uniformly sampled from \([0.5, 1.5]\) for each modality.

  \item \textbf{Intensity Adjustment:} Intensities are modified by multiplying them with a factor uniformly sampled from \([0.65, 1.2]\) with a probability of 0.15. Alternatively, intensities can be flipped using \(1 - \text{image}\). Intensity augmentation is not applied to labels. After multiplication, the values are clipped to the original intensity range.

  \item \textbf{Low Resolution:} Applied with a probability of 0.25 per sample and 0.5 per associated modality. This augmentation downsamples the triggered modalities by a factor uniformly sampled from \([1, 2]\) using nearest neighbor interpolation, then resamples them back to the original size using cubic interpolation.

  \item \textbf{Gamma Augmentation:} Applied with a probability of 0.15. Image intensities are first scaled to a range of 0 to 1, followed by a nonlinear intensity transformation defined as \(x_{\text{new}} = x_{\text{old}}^{\gamma}\), where \(\gamma\) is uniformly sampled from \([0.7, 1.5]\). The intensities are then scaled back to their original range. This augmentation is applied after the intensity flip, also with a probability of 0.15.

  \item \textbf{Spatial Flip:} Samples are flipped along all axes with a probability of 0.5.
\end{enumerate}